\documentclass{article}

 \usepackage[preprint, nonatbib]{neurips_2020}

\usepackage[utf8]{inputenc} 
\usepackage[T1]{fontenc}    
\usepackage{hyperref}       
\usepackage{url}            
\usepackage{booktabs}       
\usepackage{amsfonts}       
\usepackage{nicefrac}       
\usepackage{microtype}      

\usepackage{graphicx}
\usepackage{subcaption}
\usepackage{booktabs} 
\usepackage{amsfonts}
\usepackage{mathrsfs}
\usepackage{amsmath,amssymb, bm}
\usepackage{amsthm}
\usepackage{verbatim}
\usepackage{float}
\usepackage{multirow}

\newcommand{\eg}{\mbox{e.\,g.}\xspace}
\newcommand{\ie}{\mbox{i.\,e.}\xspace}

\newcommand*{\Scale}[2][4]{\scalebox{#1}{$#2$}}%
\DeclareMathOperator{\EX}{\mathbb{E}}

\def\bea{\begin{eqnarray}} 
	\def\eea{\end{eqnarray}}
\def\be{\begin{equation}} 
	\def\ee{\end{equation}} 
\def\ba{\begin{array}}
	\def\ea{\end{array}} 

\newtheorem{definition}{Definition}[section]

\usepackage[ruled,vlined]{algorithm2e}
\usepackage{algorithmic}

\title{
On the effect of normalization layers on Differentially Private training of deep Neural networks 
}

\author{%
  Ali Davody,\;  David Ifeoluwa Adelani,\;
Thomas Kleinbauer, and Dietrich Klakow
  \\
 Spoken Language Systems Group\\ 
  Saarland Informatics Campus\\ 
Saarland University, Germany\\
 \texttt{\{adavody|didelani|kleiba|dietrich.klakow\}@lsv.uni-saarland.de} \\
}

\begin{document}

\maketitle

\begin{abstract}
Differentially private stochastic gradient descent (DPSGD) is a variation of stochastic gradient descent based on the Differential Privacy (DP) paradigm which can mitigate privacy threats that arise from the presence of sensitive information in training data.
One major drawback of training deep neural networks with DPSGD is a reduction in the model’s accuracy. 
In this paper, we study the effect of normalization layers on the performance of DPSGD. 
We demonstrate that normalization layers have a large beneficial impact on the utility of deep neural networks with noisy parameters and should be considered essential ingredients of training
with DPSGD. In particular, we propose a novel method for integrating batch normalization with DPSGD without incurring an additional privacy loss. With our method, we are able to train deeper networks and  
achieve a better utility-privacy trade-off.  
\end{abstract}

\section{Introduction}
\label{sec:introduction}
Training deep neural networks typically requires large and representative data collections to achieve high performance. However, depending on the application domain, some datasets may contain sensitive information such as medical records of patients or personal financial data.
This has motivated the development of dedicated training methods in order to address privacy concerns (see \eg \cite{yang2017survey}).

Differential Privacy (DP) \cite{dwork2006our} provides a concrete cryptography-inspired notion of privacy. 
In practice, DP algorithms are obtained from non-private algorithms by means of appropriate randomization \cite{dwork2014algorithmic}.
Differential Privacy has been integrated into deep learning \cite{shokri2015privacy, abadi2016deep} where privacy issues can arise when the trained model permits the reconstruction of sensitive information that exists in the training data. The proposed method in \cite{abadi2016deep} is based on clipping gradients and adding random noise to them in each iteration of stochastic gradient descent (SGD). Combined with a \textit{moments accountant} method for tracing the privacy loss, this differentially  private SGD (DPSGD) technique has enabled deep neural networks to be trained under a modest privacy budget at the cost of a manageable reduction in the model's test accuracy. However, for low privacy budgets (\ie a small $\varepsilon$, see Section \ref{section:dp}), which corresponds to a large privacy guarantee, this accuracy drops significantly under DPSGD. 

The central role of input noise in the application of DPSGD deserves a more detailed analysis.
While adding a small amount of noise during training can benefit the generalization capability of the model, too much noise leads to inferior performance because of the high sensitivity of the output to perturbation in the parameters. In spite of this, we show that neural networks augmented with batch/layer normalization layers are strongly robust against random noise injection in their weights.

In this work, we investigate the impact of batch normalization \cite{ioffe2015batch} and layer normalization \cite{ba2016layer} on the performance of training under privacy constraints. Normalization layers such as batch normalization \cite{ioffe2015batch} and layer normalization \cite{ba2016layer} are indispensable components of nearly all state-of-the-art  deep neural networks.
These techniques are very essential to robustly train deep neural networks without carefully custom non-linearities \cite{klambauer2017self} or choosing a specific initialization scheme \cite{simonyan2014very}. They also improve generalization by preventing overfitting when training very deep networks \cite{zhang2019fixup}.

One important property of normalization methods is the  invariance of the model 
to weight matrix re-scaling \cite{ba2016layer}.
We argue that this invariance suggests robustness against noise injection and confirm this hypothesis empirically. In particular, we show that batch-normalization can be integrated with DPSGD without any additional loss in privacy during the training process. We compare our proposal with the current state-of-the-art methods by conducting a series of experiments on Computer Vision and Natural Language Processing tasks. 
In summary, our contributions are as follows:

\begin{itemize}
    
    \item We demonstrate that normalization layers have a substantial impact on the performance of models with noisy parameters and should be considered essential ingredients in robust differentially private training.

    \item We propose an efficient method for using batch normalization layers without incurring an additional privacy loss in the training procedure. To the best of our knowledge, our work is the first to apply a DP mechanism in the presence of batch normalization. 
    
    \item  We establish new accuracy records for differentially private trained deep networks under DPSGD on the MNIST and CIFAR10 datasets. 
    
\end{itemize}

The rest of the paper is organized as follows. We study the effect of noise on the performance of models in Section~\ref{noisy_training}. Section~\ref{section:dp} introduces our approach for differentially private training deep networks with batch normalization. We compare our method with the existing approaches in Section~\ref{section:expremints}. Finally, in section~\ref{section:related_work}, we discuss the related works  on  differential privacy.

\section{Noise and normalization}
\label{noisy_training}
In this section, we investigate how random Gaussian noise affects a network's performance in the presence of normalization layers. More specifically, we sample the weights from a Gaussian distribution $\mathcal{N}(\mu, \sigma^2)$ with learnable mean parameters $\mu$ and constant variance $\sigma^2$. Backpropagation is performed by making use of the standard reparametrization trick \cite{schulman2015gradient}.
This way of training is very similar to variational Bayesian learning of neural networks \cite{blundell2015weight}, where weights are represented by probability distributions rather than having a fixed value. Unlike the Bayesian approach though, where the goal is learning the true posterior distribution of the weights given the training data, here the noise is introduced via an ad-hoc distribution function.

Batch normalization \cite{ioffe2015batch} and layer normalization \cite{ba2016layer} are introduced to speed up deep neural network training by regularizing neuron dynamics via mean and variance statistics and reducing variance in the input to each node. Normalization techniques in combination with other architecture innovations like residual connections \cite{he2016deep} make training of very deep networks feasible.

Both batch and layer normalization ensure zero mean and unit variance in the output of a layer but using different statistics. Batch normalization (BN) calculates the mean and variance statistics across samples in a  mini-batch for each neuron independently, while layer normalization (LN) standardizes each summed input to a node utilizing the statistics over all hidden units. 

More precisely, if we denote the weighted summed inputs to the $l$-th layer by $\bm z^l = \bm{w}^{\bm{T}}\, \bm a^{l-1}$ where $a^{l-1}$ is the activation in layer $l-1$ and $\bm{w}$ is the weight matrix, then the normalization operators rescale and shift $z$ according to:

 \bea
 \tilde z_i^l =  \frac{\gamma_i}{\sqrt{\sigma_{i}^{l, 2}+\epsilon}}(z_i^l -\mu_i^l) + \beta_i
 \eea

where $\gamma_i \in \mathbb{R}$ and $ \beta_i \in \mathbb{R}$ are learnable parameters which are set to one and zero respectively at the beginning of the training. The parameters $\sigma_i^l$ and $\mu_i^l$ are estimated as follows for batch normalization:

\bea\label{bnsts}
 \mu^{\text{BN},\, l}_i = \underset{x\sim p(x)}\EX[z_i^l],\;\;\;
\sigma^{\text{BN}, l, 2}_{i} = \underset{x\sim p(x)}\EX[(z_i^l-\mu_i^l)^2],
\eea

 and as follows for layer normalization:

\bea\label{lnsts}
\mu^{\text{LN},\,l} = \frac{1}{n}\sum_{i=1}^{n}z_i^l, \;\;\;
\sigma^{\text{LN}, l} = \frac{1}{n}\sum_{i=1}^{n}(z_i^l-\mu_i^l)^2,
\eea

 where $n$ is the number of hidden units in the  layer. The expectations are estimated using samples from the training mini-batches. In the case of batch normalization, at  test time, these statistics are replaced with an exponential running average of corresponding mean and variance computed during the training phase.

In this work, we focus on the scale invariance property of normalization methods.
It is well known that both batch and layer normalization are invariant under scaling of the weight matrix (matrices) $\bm\theta \rightarrow \lambda\, \bm\theta$ with arbitrary $\lambda>0$ \cite{ba2016layer}. 
In order to demonstrate the effect of this symmetry, consider a deep neural network $f(\bm\theta^*, .)$ which characterizes the relationship from input to output
with trained parameters $\bm\theta^*$. 
The optimal weights $\bm\theta^*$ are usually learnt by minimizing a non-convex objective function, $\mathcal{L}(\bm\theta, .)$, over the training dataset using a variant of stochastic gradient descent (SGD). 
The training procedure also includes tuning hyperparameters, like learning rate and number of epochs. This tuning is usually done by maximizing the performance of the network on a validation dataset. 
The training procedure always leaves some small uncertainty of order $\delta\ll 1$ on the final values of weights. Consequently, the performance of the model with parameters  $\bm\theta^*$ and  $\bm\theta^* + \mathcal{O}(\delta)$ will be essentially  identical. For example, we may change the number of iterations or the learning rate very slightly without harming the performance of the network. 

We can make   $f(\bm\theta^*, .)$  invariant with respect to scaling of the weight matrix  by augmenting  it with batch/layer normalization operators after  each learnable layer, $f^{\text{BN/LN}}(\bm\theta^*, .)$. The re-scaling invariance implies that if the weight uncertainty in the original network is of order  $\mathcal{O}(\delta)$ then in the augmented network it can be of order  $\mathcal{O}(\lambda\delta)$ without affecting the overall performance. This suggests that neural networks with batch/layer normalization layers should be  robust against the noise in their weights. In the rest of this section, we empirically confirm this hypothesis by injecting noise into the weights of augmented networks during the training and testing procedure.


To test our hypothesis
empirically, we  train 
standard fully-connected as well as  convolutional neural networks with noise injected into their
weights on  MNIST \cite{lecun1998gradient} and  CIFAR-10  \cite{Krizhevsky09learningmultiple}.
More specifically, we sample the weights from a Gaussian distribution $\mathcal{N}(\mu, \sigma^2)$ with learnable mean parameters $\mu$ and constant variance $\sigma^2$. 
Backpropagation is performed by making use of the standard reparametrization trick \cite{schulman2015gradient}.

In particular, we investigate thoroughly LeNet-300-100 and  LeNet-5  \cite{Lecun98gradient-basedlearning} and variants of ResNet \cite{he2016deep} and VGG \cite{simonyan2014deep} models. The structure  of the models is outlined in Table~\ref{table_archs}. For each model, we construct a normalized augmented version by adding batch or layer normalization after each trainable layer.
 All models are implemented in PyTorch \cite{NEURIPS2019_9015} and trained with the Adam optimizer \cite{Kingma2014AdamAM}. 

\begin{table}[]
\centering
\resizebox{.55\columnwidth}{!}{%
\begin{tabular}{@{}l@{}c@{~~~}c@{~~}c@{~~}c@{~~}c@{~~}}
\toprule
\textit{Network}  & LeNet-5 & ResNet-18 & 
VGG \\ \midrule
\shortstack{ \textit{Convolutions}\\{}\\{}\\{}\\ {} } &
\shortstack{6, pool, 16, pool}  &
\shortstack{64, 2x[64, 64]\\2x[128, 128]\\2x[256, 256]\\2x[512, 512]} &
\shortstack{2x64 pool 2x128\\ pool, 4x256, pool} 
\\ \midrule

\textit{FC Layers} & 120, 84, 10&  avg-pool, 10 &
avg-pool, 512, 10\\ \midrule
\bottomrule
\end{tabular}
}
\vskip 0.1in
\caption{Architecture of deep networks that we use for vision tasks.}
\label{table_archs}
\end{table}

Table~\ref{table_mnist_noisy} shows  the accuracy of the augmented models on the  MNIST test dataset, averaged over ten runs against unnormalized baselines.
The BN/LN prefixes in this table denote models that are obtained by adding 
batch normalization or layer normalization layers to the original architectures, respectively. It is evident from this experiment that all augmented models are tolerant to  noise while the baselines are not. Indeed, their accuracy does not change at all for a large range of noise levels within the statistical error as promised by the scale  invariance property of the networks. 
On the other hand, the baseline models are very sensitive to small weight perturbations.  
Notably, 
disturbing the weights of the baseline models by a small noise of order $\sigma=0.3$ results in non-converging training.

\begin{table}[t]
\caption{Results of MNIST test-set accuracy ($\%\pm$ standard error) in the presence of injected noise to the weights for different models. The accuracy of models with normalization layers doesn't change within the standard deviations. In contrast, baseline models are much more sensitive to the noise, and they don't converge if the level of noise exceeds  a threshold. 
}
\label{table_mnist_noisy}
\centering
\scalebox{0.9}{
\begin{tabular}{lccccc}
        \toprule
        \multirow{3}{*}{\textbf{Model}} &  \multicolumn{5}{c}{\textbf{Noise Level ($\sigma$) }} \\
        \cmidrule(l){2-6}& 
     0 & 0.01 & 0.1 & 1 & 2  \\
        \midrule
LeNet-300-100 & $98.20\Scale[0.9]{\pm 0.07}$ &  $97.70\Scale[0.9]{\pm 0.30}$ & $ 96.98\Scale[0.9]{\pm 0.12}$ & No-convergence  & No-convergence  \\
BN-LeNet-300-100 & $98.20\Scale[0.9]{\pm 0.10}$ &  $98.10\Scale[0.9]{\pm 0.10}$ & $98.07\Scale[0.9]{\pm 0.11}$ & $98.07\Scale[0.9]{\pm 0.12}$   & $98.13\Scale[0.9]{\pm 0.08}$   \\
LN-LeNet-300-100 & $98.04\Scale[0.9]{\pm 0.16}$ &  $98.00\Scale[0.9]{\pm 0.10}$ & $98.08\Scale[0.9]{\pm 0.14}$ & $98.04\Scale[0.9]{\pm 0.09}$   & $98.03\Scale[0.9]{\pm 0.17}$  \\\midrule
LeNet-5 & $99.20\Scale[0.9]{\pm 0.02}$ & $98.94\Scale[0.9]{\pm 0.07}$ & $98.40\Scale[0.9]{\pm 0.03}$ &  No-convergence   & No-convergence  \\
BN-LeNet-5 & $99.20\Scale[0.9]{\pm 0.08}$ &  $99.21\Scale[0.9]{\pm 0.05}$ & $99.18\Scale[0.9]{\pm 0.06}$ & $99.24\Scale[0.9]{\pm 0.04}$   & $99.25\Scale[0.9]{\pm 0.07}$  \\
LN-LeNet-5 & $99.16\Scale[0.9]{\pm 0.08}$ & $99.14\Scale[0.9]{\pm 0.06}$ & $99.13\Scale[0.9]{\pm 0.07}$ & $99.21\Scale[0.9]{\pm 0.05}$   & $99.19\Scale[0.9]{\pm 0.07}$ 
\\\bottomrule
\end{tabular}
}
\end{table}

\begin{table}[t]
\caption{CIFAR-10 test-set percentage accuracy ($\%\pm$ standard error) with noisy weights for a variety of models. We see the same pattern as in MNIST dataset where models augmented with batch normalization are very robust against noise.}
\label{table_cifar_noisy}
\centering
\scalebox{0.95}{
\begin{tabular}{lcccccc}
\toprule
        \multirow{3}{*}{\textbf{Model}} &  \multicolumn{5}{c}{\textbf{Noise Level ($\sigma$) }} \\
        \cmidrule(l){2-6}& 
     0 & 0.01 & 0.1 & 1 & 2  \\
        \midrule
ResNet-18 & $93.50\Scale[0.9]{\pm 0.04}$ &  $89.05\Scale[0.9]{\pm 0.60}$ & $89.41\Scale[0.9]{\pm 0.67}$ & $86.76\Scale[0.9]{\pm 1.30}$   & $87.83\Scale[0.9]{\pm 0.87}$  \\
ResNet-18-mod & $93.60\Scale[0.9]{\pm 0.18}$ &  $88.65\Scale[0.9]{\pm 1.22}$ & $88.16\Scale[0.9]{\pm 1.22}$ & $85.46\Scale[0.9]{\pm 1.21}$   & $83.95\Scale[0.9]{\pm 2.51}$   \\
BN-ResNet-18 & $93.55\Scale[0.9]{\pm 0.04}$ &  $93.59\Scale[0.9]{\pm 0.18}$ & $93.46\Scale[0.9]{\pm 0.24}$ & $92.72\Scale[0.9]{\pm 0.26}$   & $92.10\Scale[0.9]{\pm 0.28}$  \\\midrule
VGG-16 & $91.21\Scale[0.9]{\pm 0.16}$ &  $88.89\Scale[0.9]{\pm 0.56}$ & $84.59\Scale[0.9]{\pm 6.11}$ & $63.12\Scale[0.9]{\pm 8.21}$   & No-convergence \\
VGG-16-mod & $91.18\Scale[0.9]{\pm 0.20}$ &  $88.87\Scale[0.9]{\pm 0.59}$ & $82.58\Scale[0.9]{\pm 6.91}$ & $22.58\Scale[0.9]{\pm 5.0}$   & No-convergence   \\
BN-VGG-16 & $91.75\Scale[0.9]{\pm 0.22}$ &  $91.69\Scale[0.9]{\pm 0.20}$ & $91.56\Scale[0.9]{\pm 0.24}$ & $90.95\Scale[0.9]{\pm 0.25}$   & $90.35\Scale[0.9]{\pm 0.14}$ 
\\\bottomrule
\end{tabular}
}
\vskip -0.1in
\end{table}

Experiments on CIFAR-10 with ResNet and VGG networks show similar trends (see Table~\ref{table_cifar_noisy}). Unlike LeNet models, the original ResNet and VGG networks already contain BN layers after all except the last trainable layer. This leads to some degree of protection against noise as demonstrated in Table~\ref{table_cifar_noisy}. For example, the accuracy of ResNet-18 with a noise level of $1$ reduces only to $87\%$ instead of to random prediction. 

To illustrate the role of normalization layers further, we also present results on ResNet-18-mod  and VGG-16-mod  which are obtained by removing the last normalization layer from the original architectures. As shown by the performance degradation with increasing noise levels in Table~\ref{table_cifar_noisy}, these models are more vulnerable to the added noise.

Next, we extended our experiments to a more complex task, viz. natural language text classification, where we observe a similar effect. For this, we trained a BiLSTM model with one linear/dense layer (DL) with/without layer normalization
on the AG News Corpus\footnote{\url{http://groups.di.unipi.it/~gulli/AG_corpus_of_news_articles.html}}, a popular text classification dataset with four categories of news: \textit{World}, \textit{Sports}, \textit{Business} and \textit{Sci/Tech}. 
Each class has 30,000 training examples and 1,900 test examples. In total, the dataset consists of 120,000 training examples and 7,600 test examples. We further split the training examples into training/validation set where we use 96,000 examples for training the models and 24,000 for validation (e.g early stopping and for tuning adaptive learning rate). 
We trained the models with different noise levels $\sigma$ for 25 epochs. We find that the higher the noise level, the more 
epochs are needed to maintain the accuracy of the baseline model. Our results on the language data in Table~\ref{Tab:table_ag_noisy} suggest that layer normalization makes the BiLSTM model robust to noise with minimal drops in accuracy ($1-7\%$). 
Our findings make us conclude that the LSTM and CNN architectures are robust to noise when equipped with layer and batch normalization.

\begin{figure}[ht]
     \centering
     \begin{subfigure}{0.45\textwidth}
         \centering
         \includegraphics[scale=.25]{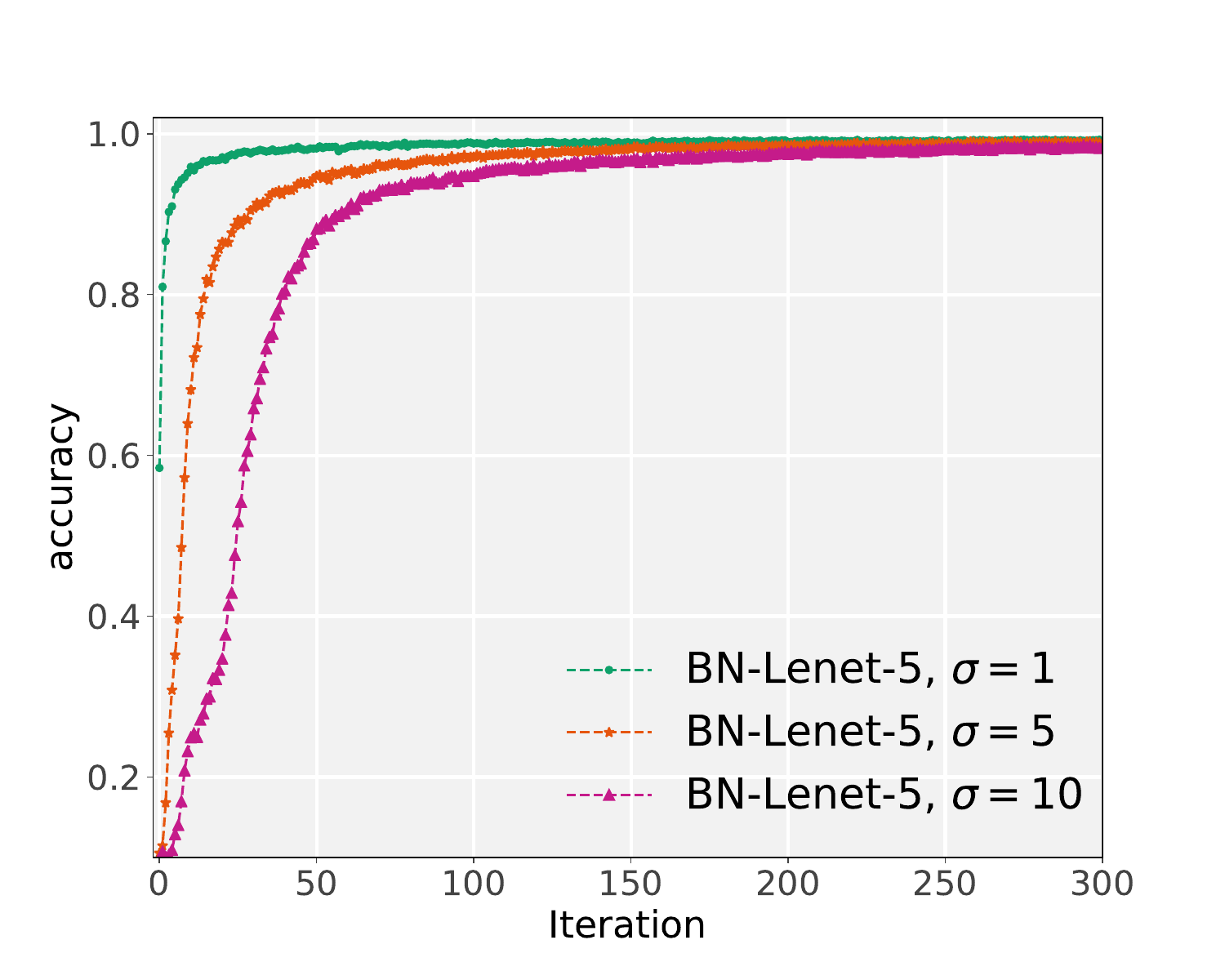}
         \caption{MNIST}
     \end{subfigure}%
     ~ 
     \begin{subfigure}{0.45\textwidth}
         \centering
         \includegraphics[scale=.25]{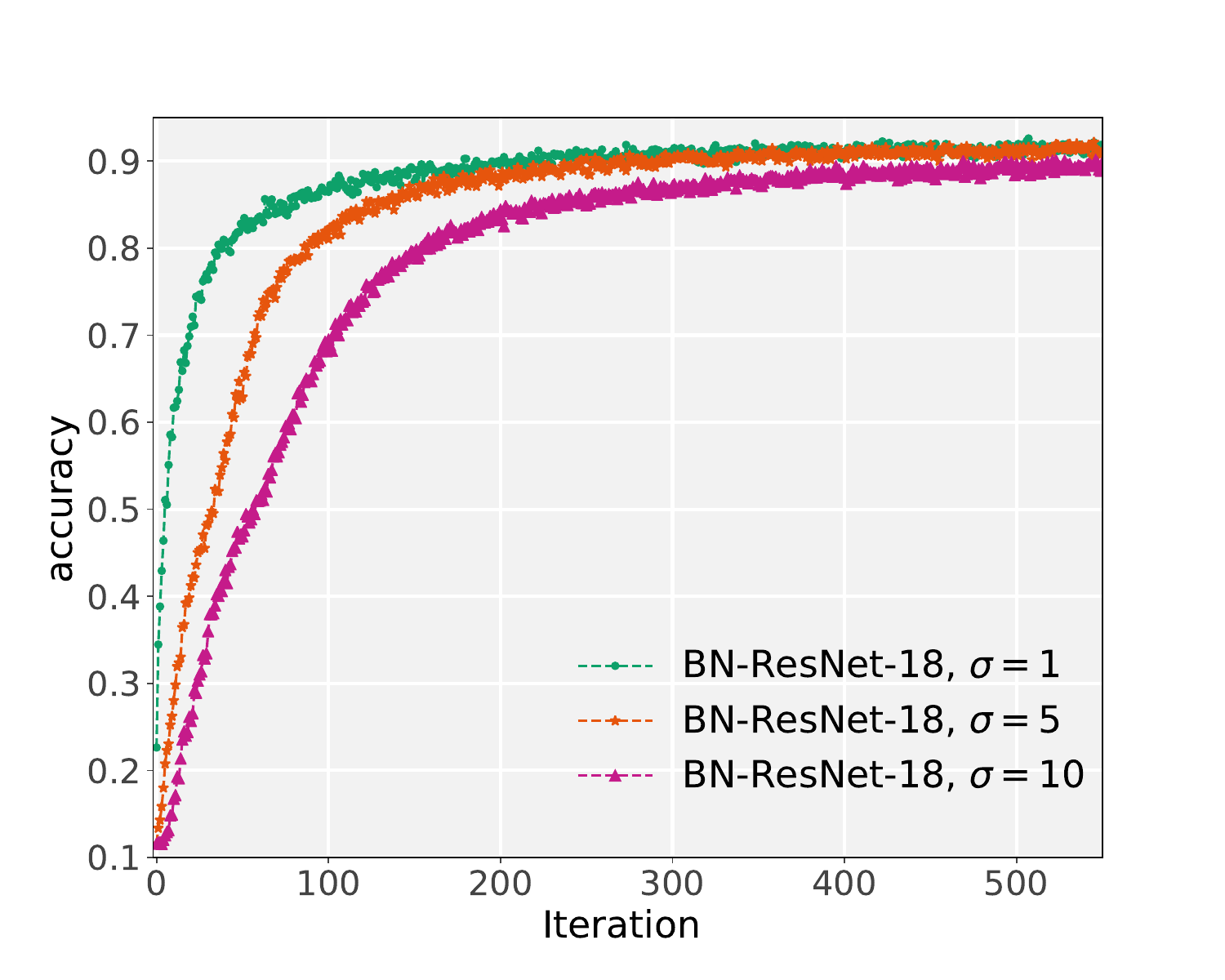}
         \caption{CIFAR-10}
     \end{subfigure}
     \caption{Evolution of validation accuracy during training for MNIST and CIFAR-10 datasets. A large value of noise slows down the training but does not affect the performance drastically. }
\label{plot_acc}
\end{figure}

It is worth mentioning that achieving the same accuracy in the presence of noise does not come for free as it affects the training time: the larger the noise, the slower the training. Figure~\ref{plot_acc}  illustrates the evolution of the accuracy of models on the validation set, for different levels of noise. As it is evident from these plots, models with different values of noise converge to the same  accuracy, albeit with different rates.
For example, increasing the noise level from 1 to 10 slows down training by a factor of order 7 for LeNet and ResNet models. More details on this can be found in Appendix~2.

\begin{table}[t]
\caption{AG News Corpus test accuracy with noisy weights for a variety of models. We see the same pattern as the vision dataset where models augmented with layer normalization are very robust against noise.}
\label{Tab:table_ag_noisy}
\centering
\scalebox{0.95}{
\begin{tabular}{lcccccc}
        \toprule    
         \multirow{3}{*}{\textbf{Model}} &  \multicolumn{5}{c}{\textbf{Noise Level ($\sigma$) }} \\
 \cmidrule(l){2-6}& 
     0 & 0.01 & 0.1 & 1 & 2  \\
         \midrule
BiLSTM-DL & $89.34\%$ &  $89.57\%$ & $89.01\%$ & $66.32\%$   & $24.76\%$  \\
LN-BiLSTM-DL & $89.34\%$ &  $88.87\%$ & $88.62\%$ & $85.74\%$   & $82.41\%$ 
\\\bottomrule
\end{tabular}
}
\end{table}

\section{DPSGD and Normalization}
\label{section:dp}
Differential privacy (DP) is a systematic approach to
quantifying  a privacy guarantee while querying a dataset.
quantifying the leakage of information due to query a dataset. DP protects privacy by bounding the influence of any sample on the outcome of queries. It provides a provable guarantee for individuals.
 We proceed by briefly recalling  some preliminaries on differential privacy and then propose our approach for training deep neural networks in a differentially private way.

Let us denote the domain of data points by $\chi$. We call two datasets $D_1, D_2 \in \chi$ \emph{ neighboring} if they differ 
are the same except one data point, 
\ie, $d(D_1, D_2)=1$, where $d(., .)$ is the Hamming distance.  

\begin{definition}{(Differential Privacy \cite{dwork2006calibrating}).}
A randomized algorithm $\mathcal{M}: \chi \rightarrow \mathcal{R}$ with domain $\chi$ and range $\mathcal{R}$ is $(\varepsilon, \delta)$ differential private if for all measurable sets $S \in \mathcal{R}$
and for all neighboring datasets $D_1$ and $D_2$, 
it holds that
\vspace{-5mm}

\begin{equation}
    \Pr[\mathcal{M}(D_1) \in S] \leq \exp{(\varepsilon)} \Pr[\mathcal{M}(D_2) \in S] + \delta.
\end{equation}

\end{definition}

Intuitively a $(\varepsilon, \delta)$ differential private mechanism guarantees that the absolute value of privacy leakage will be bounded by $\varepsilon$ with probability at least  $1-\delta$ for adjacent datasets. The higher the value of $\varepsilon$, the more the chance of data re-identification and so information leakage. 

A standard approach for achieving differential privacy is to add some random noise $r$ to the output of queries, $q(D) + r$, and to tune the noise $r$ by the \textit{sensitivity} of the query. $L_p$ sensitivity is defined  as  the maximum change in the outcome of a query for two neighboring datasets and measures the maximum influence that a single data point can have on the result of the query:

 \bea
 S_p = \max_{d(D_1, D_2)=1} \| q(D_2) -q(D_1)\|_p.
 \eea

The special case where $r$ is calibrated with $S_2$ sensitivity and sampled according to the normal distribution is of special importance and is termed \textit{Gaussian mechanism}:

\be
G(D)\colon = q(D) + \mathcal{N}(0, S_2^2 \sigma^2).
\ee
Here $\mathcal{N}(0, S_2^2 \sigma^2)$ is the normal distribution with mean zero and standard deviation $S_2 \sigma$. It can be shown that this mechanism satisfies $(\varepsilon, \delta)$ differential privacy provided that $\sigma \ge \frac{\sqrt{2\ln{\frac{1.25}{\delta}}}}{\varepsilon}$ \cite{dwork2014algorithmic}. 

Differential privacy has been integrated into deep learning in \cite{shokri2015privacy} and subsequently in  \cite{abadi2016deep}
for the setting where an adversary has access to the network architecture and learned weights, $f(\bm{\theta}^*, .)$. In particular, the method in \cite{abadi2016deep} preserves privacy by adding noise to the SGD updates:

\vspace{-3mm}
\be\label{dpsgd_update}
\bm\theta_{t+1} \leftarrow \bm\theta_t -\eta\,\mathbf{g}_t + \frac{\eta}{L} r, 
\ee
\vspace{-3mm}

where $\mathbf{g}_t$ is the averaged gradient, $\eta$ is the learning rate and $r$ is sampled from the Gaussian distribution $\mathcal{N}(0, \sigma^2)$. To control the influence of training samples on the parameters, the gradients are clipped by the $L_2$-norm. 
 \be\label{clipping}
 \pi(g_i) =g_i \,. \min(1, C/\|g_i\|_2),
 \ee
where $g_i$ is the gradient corresponding to the $i$-th  sample and $C$ is the clipping factor.
It has been shown in \cite{abadi2016deep}  that each step of DPSGD is $(\varepsilon, \delta)$-differential private once we tune the noise as $\sigma=C\, z$ with $z =  \frac{\sqrt{2\ln{\frac{1.25}{\delta}}}}{\varepsilon}$.

It is known that batch normalization is not consistent with DP training.  
Indeed, in a non-private setting, one usually keeps track of the running averages of mean and variance statistics (Eq.~\ref{bnsts})
during the training procedure and reuses this collected information at test time to normalize the inputs to neurons.
 More specifically, the update rule for running averages at each iteration is as follows:

\bea
&&\bm\mu^{\text{Batch}} \leftarrow (1-\alpha)\bm\mu^{\text{Batch}} + \alpha \,\bm\mu^{\text{Batch}}_{t}\,, \\
&&\bm\sigma^{2, \text{Batch}} \leftarrow (1-\alpha) \bm\sigma^{2, \text{Batch}} + \alpha \, \bm\sigma^{2, \text{Batch}}_{t},
\eea

where ($\bm\mu^{\text{Batch}}$, $\bm\sigma^{2, \text{Batch}}$) are estimated mean and variance statistics, ($\bm\mu^{\text{Batch}}_t$, $\bm\sigma^{2, \text{Batch}}_t$) are new observed values on iteration $t$ of training according to Eq.~\ref{bnsts} and $\alpha$ is the momentum of moving averages.

Since these running averages are also a part of the model's outputs, 
in a private training,  
we have to add noise also to these statistics at each iteration and distribute the privacy budgets among the weights and moving averages to make the overall procedure differentially private.
Additionally, we need to truncate the summed inputs of neurons to bound the sensitivity of means and variances, which are given by \cite{swanberg2019improved}:

\bea\nonumber
&&S_\mu = \frac{C'}{L}, \\
&&S_{\sigma^2} = C'^2\left(\frac{3}{L} - \frac{3}{L^2} +\frac{1}{L^3}\right),
\nonumber
\eea

where $C'$ is the clipping threshold for neurons activations and $L$ is the batch size. Empirically, we have found that if we tune the noise with the worst-case scenario according to the above sensitivities, the performance of the model drops drastically. Therefore we employ a more sophisticated approach to deal with batch normalization situation as shown in algorithm~\ref{alg:SI-DPSGDbn}.

First of all,  we do not track the running averages during the training phase and instead computed fresh statistics  from the current batch are used to normalize the neurons. But to be able to deal with batch normalization at test time we concatenate a fixed amount of data points $\hat{X}$ with size $M$ taken from a public dataset, disjoint from the training data, to the input of the network, both in the training and the test phase. These samples only contribute to the statistics and not to the cost function directly.

\begin{minipage}[]{\textwidth}
\begin{algorithm}[H]
\caption{{\sc DPSGD with Batch Normalization: Training}}
\label{alg:SI-DPSGDbn}
\begin{algorithmic}
\STATE {\bfseries Input:} dataset $\mathcal{D}=\{(x_1, y_1), \cdots\}$ of size $N$, 
 	 a public dataset $\hat X = \{\hat x_1, \hat x_2, \cdots, \hat x_{M}\}$, loss function $\mathcal{L}(\bm\theta, .)$, learning rate $\eta_t$, noise multiplier $z$, sample size $L$, gradient norm bound $C$ and  $T$ iterations. 
\vspace{1mm}
\FOR{$t=0$ {\bfseries to} $T-1$}
		\STATE $\bullet$ Take  a random sample, $X\sim \mathcal{D}$, with size $L$ and selection probability $\frac{L}{N}$.

        \STATE $\bullet$	Concatenate the public data to each lot\\
		~~~ $X \leftarrow X \cup \hat X $ \\

        \STATE $\bullet$ Compute lost for the first $L$ elements\\
 		~~~ $\mathcal{L}(\bm\theta_t, x_i) =  \mathcal{L}(\bm\theta_t, X)[i]$\\

		\STATE	$\bullet$ Compute gradient  \\
		~~$\;\;\mathbf{g}_t(x_i) \leftarrow  \nabla_{\bm{\mu}_t}  \mathcal{L}(\bm\theta_t(\bm\mu_t), x_i)$.
					
		\STATE $\bullet$ Clip gradient \\
 		~~~~$\mathbf{g}_t(x_i) \leftarrow \mathbf{g}_t(x_i) \; .\; \min(1, C/\Vert\mathbf{g}_t(x_i)\Vert_2)$\\
		\STATE	$\bullet$	Add noise \\
		~~~~$\mathbf{g}_t \leftarrow \frac{1}{L}\big( \sum_{i} \mathbf{g}_t(x_i) +  \mathcal{N}(0, C^2 z^2)\big)$\\
					
		\STATE	 $\bullet$ 	Update  parameters:\\
 		~~~~$\bm{\theta}_{t+1} \leftarrow \bm{\theta}_{t}  - \eta_t\;  \mathbf{g}_t$
\\\hrulefill
\vspace{1.mm}
\STATE  \textbf{Test Phase}
\vspace{-1.5mm}
\\\hrulefill
\STATE {\bfseries Input:} test dataset $\mathcal{D}_{test}=\{(x_1, y_1), \cdots\}$ of size $N_{test}$, 
the public dataset $\hat X = \{\hat x_1, \hat x_2, \cdots, \hat x_{M}\}$, trained model $f(\bm{\theta}_{T}, .)$.
\vspace{2mm}

\STATE $\bullet$ 	Initialize Y $\leftarrow \emptyset$\\
\FOR{$i=0$ {\bfseries to} $N_{test}-1$}

\STATE $\bullet$ 	Concatenate the  public dataset to each data point $x_i$\\
~~~~$X \leftarrow x_i \cup \hat X $ 

\STATE $\bullet$  Compute the network output corresponding to the data point $x_i$\\
~~~~$y_i =  f(\bm{\theta}_{T}, X)[0]$\\
\STATE $\bullet$Append $y_i$ to the results\\
~~~~$Y \leftarrow Y \cup y_i$

\ENDFOR

\STATE {\bfseries Return} outputs of network $Y$.

\ENDFOR
 \end{algorithmic}
\end{algorithm}
\end{minipage}

%
%
%

Therefore, in the training phase, the cost is computed via $\mathcal{L}(\bm\theta_t, X \cup \hat X)[:L]$, where $X$ is a batch of size $L$ from the training data, and $[:L]$ denotes the slice of the first $L$ elements. At  test time, when iterating over the dataset, the same public data points $\hat{X}$ are also concatenated to each test sample, $x$, and the output of network is computed as $f(\bm{\theta}_{T}, x \cup \hat X)[0]$. This leads to privacy preserving batch normalization and allows us to compute the normalization statistics over a batch of size $1+M$ without any reference to the training data.

In the next section, we present the impact of normalization layers on DP training using two image recognition  benchmarks, i.e. MNIST and CIFAR-10,  as well as text classification task in natural language processing using AG News Corpus.

\section{Experiments}
\label{section:expremints}

In this section, we report some results of applying our method and compare them with  existing DP mechanisms. 
The purpose of these experiments is two-fold: 
we show that (1) similar to non-private training (section \ref{noisy_training}), normalization layers improve the performance of models trained with DPSGD; and
(2) that training very deep networks is feasible with our private version of batch normalization.

All models, as well as DPSGD, have been implemented in PyTorch \cite{paszke2017automatic}. To track the privacy loss over the whole training procedure, we employ the Rényi-DP technique  \cite{mironov2017renyi}.
It provides a tighter bound on the privacy loss compared to the strong composition theorem \cite{dwork2010boosting}.
We use the open-source implementation of the Rényi DP accountant from the TensorFlow Privacy package
\cite{TFP} 
 \footnote{\url{https://github.com/tensorflow/privacy}}. 
The total privacy loss  $\varepsilon$ is computed as a function of the  noise multiplier $z$, size of dataset $N$, size of lot $L$, the number of iterations $T$ and $\delta$.

Table~\ref{t: mnist_dp} depicts the accuracy of LeNet-5 model on the MNIST test set for $\varepsilon$ ranging from high to very low privacy budgets. We have trained LeNet-5 with and without normalization layers, using DPSGD.
For training the model augmented with batch normalization, we employed 
128 images of KMNIST \cite{clanuwat2018deep} as the public dataset. The probability $\delta$ parameter is set to $10^{-5}$ in all our experiments.

The results illustrate that the use of normalization layers consistently improves  the performance of DPSGD for all finite values of privacy loss. Further, we observe that the effect of batch normalization is greater than that of layer normalization.
Remarkably,  we gain around $7\%$ and $10\%$  for very low privacy budgets of $\varepsilon=0.1$ and $\varepsilon=0.05$ with our private batch normalization technique.

We now turn our attention to the CIFAR-10 dataset.
Table~\ref{t: cifar_dp} summarizes the results of DPSGD on the TensorFlow tutorial model considered in \cite{abadi2016deep}. 
We follow the same experimental setting as in \cite{abadi2016deep}, i.e. we fine tune the linear layers of a pretrained model trained on the CIFAR-100 dataset. 
For training the model with batch normalization we also use 128 images of CIFAR-100%
, which has completely  different image examples and classes from those of CIFAR-10,
 as our public dataset.

As Table~\ref{t: cifar_dp} illustrates, 
using batch normalization results in better accuracies than layer normalization as well as raw  TensorFlow tutorial model. We also show the results of training a light weight VGG model in this table (See appendix A for the details of architecture). The non-private accuracy of this model is comparable with the TF-tutorial model but it leads to a much lower gap for finite privacy budgets. It should be mentioned that it is not feasible to train such models without making use of the batch normalization as the training is extremely unstable. Therefore our private friendly batch normalization technique allows training much more deeper and complex networks and establishing new scores for the performance of differentially private models.

Next, we extended our experiments to a more complex task, viz. natural language text classification, where we observe a similar effect. For this, we trained a BiLSTM model with one linear/dense layer (DL) with/without layer normalization (LN) on the AG News Corpus~\footnote{\url{http://groups.di.unipi.it/~gulli/AG_corpus_of_news_articles.html}}, a popular text classification dataset with 4 categories of news: \textit{World}, \textit{Sports}, \textit{Business} and \textit{Sci/Tech}. Table~\ref{Tab: ag_dp} shows the result of our experiment on text classification. As before, layer normalization has a large impact on the performance of the model.

\begin{table*}[]
  \centering
  \footnotesize
  \caption{Test accuracy of three DP training methods on  MNIST with different privacy loss ($\delta=10^{-5}$). 
  }
\vspace{0mm}
\label{t: mnist_dp}
\scalebox{.95}{
    \begin{tabular}{lccccccccc}
    \toprule
 \multirow{3}{*}{\textbf{DP Algorithm}} &  \multicolumn{7}{c}{\textbf{privacy budget ($\varepsilon$)}} \\
 \cmidrule(l){2-8} & 
     $\infty$ & 7 & 3 & 1 & 0.5 & 0.1 & 0.05  \\
 \midrule
DPSGD (LeNet-5) & $99.20\%$ & $97.01\%$ &  $96.34\%$ & $94.11\%$ & $91.10\%$  & $83.00\%$ & $78.96\%$  \\
\addlinespace[0.3em]
DPSGD (LN-LeNet-5) & $99.20\%$ & $97.35\%$ &  $97.05\%$ & $96.68\%$ & $94.81\%$  & $87.45\%$ & $75.76\%$  \\
\addlinespace[0.3em]
DPSGD (BN-LeNet-5) & $99.20\%$ & $\bm{98.68}\%$ &  $\bm{98.18}\%$ & $\bm{97.61\%}$ & $\bm{96.83\%}$ & $\bm{90.68\%}$ & $\bm{88.15}\%$   \\ 
     \bottomrule
    \end{tabular}
    }
\end{table*}

\begin{table*}[]
  \centering
  \footnotesize
  \caption{Results on CIFAR-10 test-set accuracy with $\delta=10^{-5}$.
  }
 \vspace{0mm}
\label{t: cifar_dp}
\scalebox{0.9}{
    \begin{tabular}{lcccccc}
    \toprule
 \multirow{3}{*}{\textbf{DP Algorithm}} &  \multicolumn{4}{c}{\textbf{privacy budget ($\varepsilon$)}} \\
 \cmidrule(l){2-5}& 
     $\infty$  & $8$  & $4$  &  $2$  \\
 \midrule
DPSGD (TF-tutorial) \cite{abadi2016deep} & $80.0\%$  & $73.0\%$ &  $70.0\%$ & $67.0\%$ \\ \addlinespace[0.3em]
DPSGD (LN-TF-tutorial) & $80.0\%$ & $73.3
\%$ &  $70.6\%$ & $67.0\%$  \\
\addlinespace[0.3em]
DPSGD(BN-TF-tutorial) & $80.0\%$ & $\bm{74.1\%}$ &  
$\bm{71.2\%}$ & $\bm{69.8\%}$ \\ 
\midrule
\addlinespace[0.3em]
DPSGD (BN-VGG) & $80.7
\%$ & $\bm{79.5}\%$ &  $\bm{79.1}\%$ & $\bm{77.4}\%$  \\  
\bottomrule
\end{tabular}
}    
\end{table*}

\begin{table*}[h]
  \centering
  \footnotesize
  \caption{Testing accuracy of  differentially private training methods on  AG News Corpus with different privacy loss and $\delta=10^{-5}$. 
  }
  \vspace{2mm}
\label{Tab: ag_dp}
\scalebox{0.85}{
    \begin{tabular}{lcccccccc}
    \toprule
 \multirow{3}{*}{\textbf{DP Algorithm}} &  \multicolumn{6}{c}{\textbf{privacy budget ($\varepsilon$)}} \\
 \cmidrule(l){2-7}& 
     $\infty$ & 7 & 3 & 1 & 0.5 & 0.1   \\
 \midrule
DPSGD (BiLSTM-DL) & $88.47\%$ & $83.86\%$ &  $80.00\%$ & $81.14\%$ & $77.88\%$  & $37.49\%$  \\ \addlinespace[0.3em]
 DPSGD (LN-BiLSTM-DL) & $88.18\%$ & $\bm{84.34\%}$ &  $\bm{82.51\%}$ & $\bm{82.03\%}$ & $\bm{79.16\%}$  & $\bm{50.09\%}$   \\\addlinespace[0.3em]
     \bottomrule
    \end{tabular}
}    
  \end{table*}

\section{Related Work}
\label{section:related_work}
A number of different methods have been developed to preserve privacy in machine learning models. \cite{shokri2015privacy} proposed a distributed multiparty learning mechanism for a  network without sharing input datasets, however, the obtained privacy guarantee was very loose. \cite{abadi2016deep} developed an efficient differentially private SGD for training networks with a large number of parameters. Also, one may employ tighter bounds provided by the Rényi differential privacy \cite{8049725} in conjunction with DPSGD.
A method for adding less noise to the weights of neural networks is proposed in 
\cite{pichapati2019adaclip} by using  adaptive clipping of the gradients. \cite{papernot2020making} shows
that learning with DPSGD requires optimization of the model architectures and initializations.
Perturbing the objective function as an alternative way to protect privacy has been suggested in \cite{phan2016differential}.
A model based on the local differentially private mechanism is proposed in \cite{chamikara2019local}  to train deep convolutional networks in a way that a data owner can add a randomization layer before data leave data owners’ device. 
A comprehensive list of works in this area
can be found in \cite{zhu2017differential}.

\section{Conclusion}
In this paper, we proposed a novel method for integrating batch normalization with differentially private stochastic gradient descent. Our method makes training very deep neural networks, such as ResNet and VGG, feasible under very strong privacy guarantees. We also have demonstrated that normalization layers are  essential ingredients for a robust private training.

\section{Acknowledgments}
We would like to thank 
Marius Mosbach and Xiaoyu Shen for proof-reading 
and 
valuable comments. 
The presented research has been funded by the European Union's Horizon 2020 research and innovation programme project COMPRISE (\texttt{http://www.compriseh2020.eu/}) under grant agreement No. 3081705.

\bibliography{main}

\begin{thebibliography}{10}

\bibitem{abadi2016deep}
M.~Abadi, A.~Chu, I.~Goodfellow, H.~B. McMahan, I.~Mironov, K.~Talwar, and
  L.~Zhang.
\newblock Deep learning with differential privacy.
\newblock In {\em Proceedings of the 2016 ACM SIGSAC Conference on Computer and
  Communications Security}, pages 308--318. ACM, 2016.

\bibitem{ba2016layer}
J.~L. Ba, J.~R. Kiros, and G.~E. Hinton.
\newblock Layer normalization.
\newblock {\em arXiv preprint arXiv:1607.06450}, 2016.

\bibitem{blundell2015weight}
C.~Blundell, J.~Cornebise, K.~Kavukcuoglu, and D.~Wierstra.
\newblock Weight uncertainty in neural networks.
\newblock {\em arXiv preprint arXiv:1505.05424}, 2015.

\bibitem{chamikara2019local}
M.~Chamikara, P.~Bertok, I.~Khalil, D.~Liu, and S.~Camtepe.
\newblock Local differential privacy for deep learning.
\newblock {\em arXiv preprint arXiv:1908.02997}, 2019.

\bibitem{clanuwat2018deep}
T.~Clanuwat, M.~Bober{-}Irizar, A.~Kitamoto, A.~Lamb, K.~Yamamoto, and D.~Ha.
\newblock Deep learning for classical japanese literature.
\newblock {\em CoRR}, abs/1812.01718, 2018.

\bibitem{dwork2006our}
C.~Dwork, K.~Kenthapadi, F.~McSherry, I.~Mironov, and M.~Naor.
\newblock Our data, ourselves: Privacy via distributed noise generation.
\newblock In {\em Annual International Conference on the Theory and
  Applications of Cryptographic Techniques}, pages 486--503. Springer, 2006.

\bibitem{dwork2006calibrating}
C.~Dwork, F.~McSherry, K.~Nissim, and A.~Smith.
\newblock Calibrating noise to sensitivity in private data analysis.
\newblock In {\em Theory of cryptography conference}, pages 265--284. Springer,
  2006.

\bibitem{dwork2014algorithmic}
C.~Dwork, A.~Roth, et~al.
\newblock The algorithmic foundations of differential privacy.
\newblock {\em Foundations and Trends{\textregistered} in Theoretical Computer
  Science}, 9(3--4):211--407, 2014.

\bibitem{dwork2010boosting}
C.~Dwork, G.~N. Rothblum, and S.~Vadhan.
\newblock Boosting and differential privacy.
\newblock In {\em 2010 IEEE 51st Annual Symposium on Foundations of Computer
  Science}, pages 51--60. IEEE, 2010.

\bibitem{he2016deep}
K.~He, X.~Zhang, S.~Ren, and J.~Sun.
\newblock Deep residual learning for image recognition.
\newblock In {\em Proceedings of the IEEE conference on computer vision and
  pattern recognition}, pages 770--778, 2016.

\bibitem{ioffe2015batch}
S.~Ioffe and C.~Szegedy.
\newblock Batch normalization: Accelerating deep network training by reducing
  internal covariate shift.
\newblock {\em arXiv preprint arXiv:1502.03167}, 2015.

\bibitem{Kingma2014AdamAM}
D.~P. Kingma and J.~Ba.
\newblock Adam: A method for stochastic optimization.
\newblock {\em CoRR}, abs/1412.6980, 2014.

\bibitem{klambauer2017self}
G.~Klambauer, T.~Unterthiner, A.~Mayr, and S.~Hochreiter.
\newblock Self-normalizing neural networks.
\newblock In {\em Advances in neural information processing systems}, pages
  971--980, 2017.

\bibitem{Krizhevsky09learningmultiple}
A.~Krizhevsky.
\newblock Learning multiple layers of features from tiny images.
\newblock Technical report, CIFAR, 2009.

\bibitem{Lecun98gradient-basedlearning}
Y.~LeCun, L.~Bottou, Y.~Bengio, and P.~Haffner.
\newblock Gradient-based learning applied to document recognition.
\newblock In {\em Proceedings of the IEEE}, pages 2278--2324, 1998.

\bibitem{lecun1998gradient}
Y.~LeCun, L.~Bottou, Y.~Bengio, P.~Haffner, et~al.
\newblock Gradient-based learning applied to document recognition.
\newblock {\em Proceedings of the IEEE}, 86(11):2278--2324, 1998.

\bibitem{mironov2017renyi}
I.~Mironov.
\newblock R{\'e}nyi differential privacy.
\newblock In {\em 2017 IEEE 30th Computer Security Foundations Symposium
  (CSF)}, pages 263--275. IEEE, 2017.

\bibitem{8049725}
I.~{Mironov}.
\newblock Rényi differential privacy.
\newblock In {\em 2017 IEEE 30th Computer Security Foundations Symposium
  (CSF)}, pages 263--275, Aug 2017.

\bibitem{papernot2020making}
N.~Papernot, S.~Chien, S.~Song, A.~Thakurta, and U.~Erlingsson.
\newblock Making the shoe fit: Architectures, initializations, and tuning for
  learning with privacy, 2020.

\bibitem{paszke2017automatic}
A.~Paszke, S.~Gross, S.~Chintala, G.~Chanan, E.~Yang, Z.~DeVito, Z.~Lin,
  A.~Desmaison, L.~Antiga, and A.~Lerer.
\newblock Automatic differentiation in pytorch.
\newblock {\em nips}, 2017.

\bibitem{NEURIPS2019_9015}
A.~Paszke, S.~Gross, F.~Massa, A.~Lerer, J.~Bradbury, G.~Chanan, T.~Killeen,
  Z.~Lin, N.~Gimelshein, L.~Antiga, A.~Desmaison, A.~Kopf, E.~Yang, Z.~DeVito,
  M.~Raison, A.~Tejani, S.~Chilamkurthy, B.~Steiner, L.~Fang, J.~Bai, and
  S.~Chintala.
\newblock Pytorch: An imperative style, high-performance deep learning library.
\newblock In H.~Wallach, H.~Larochelle, A.~Beygelzimer, F.~d~Alch\'{e}-Buc,
  E.~Fox, and R.~Garnett, editors, {\em Advances in Neural Information
  Processing Systems 32}, pages 8024--8035. Curran Associates, Inc., 2019.

\bibitem{phan2016differential}
N.~Phan, Y.~Wang, X.~Wu, and D.~Dou.
\newblock Differential privacy preservation for deep auto-encoders: an
  application of human behavior prediction.
\newblock In {\em Thirtieth AAAI Conference on Artificial Intelligence}, 2016.

\bibitem{pichapati2019adaclip}
V.~Pichapati, A.~T. Suresh, F.~X. Yu, S.~J. Reddi, and S.~Kumar.
\newblock Adaclip: Adaptive clipping for private sgd, 2019.

\bibitem{schulman2015gradient}
J.~Schulman, N.~Heess, T.~Weber, and P.~Abbeel.
\newblock Gradient estimation using stochastic computation graphs.
\newblock In {\em Advances in Neural Information Processing Systems}, pages
  3528--3536, 2015.

\bibitem{shokri2015privacy}
R.~Shokri and V.~Shmatikov.
\newblock Privacy-preserving deep learning.
\newblock In {\em Proceedings of the 22nd ACM SIGSAC conference on computer and
  communications security}, pages 1310--1321. ACM, 2015.

\bibitem{simonyan2014very}
K.~Simonyan and A.~Zisserman.
\newblock Very deep convolutional networks for large-scale image recognition.
\newblock {\em arXiv preprint arXiv:1409.1556}, 2014.

\bibitem{simonyan2014deep}
K.~Simonyan and A.~Zisserman.
\newblock Very deep convolutional networks for large-scale image recognition,
  2014.

\bibitem{swanberg2019improved}
M.~Swanberg, I.~Globus-Harris, I.~Griffith, A.~Ritz, A.~Groce, and A.~Bray.
\newblock Improved differentially private analysis of variance.
\newblock {\em arXiv preprint arXiv:1903.00534}, 2019.

\bibitem{TFP}
\url{https://github.com/tensorflow/privacy}.
\newblock Accessed: 2020-January.

\bibitem{yang2017survey}
Y.~Yang, L.~Wu, G.~Yin, L.~Li, and H.~Zhao.
\newblock A survey on security and privacy issues in internet-of-things.
\newblock {\em IEEE Internet of Things Journal}, 4(5):1250--1258, 2017.

\bibitem{zhang2019fixup}
H.~Zhang, Y.~N. Dauphin, and T.~Ma.
\newblock Fixup initialization: Residual learning without normalization.
\newblock {\em arXiv preprint arXiv:1901.09321}, 2019.

\bibitem{zhu2017differential}
T.~Zhu, G.~Li, W.~Zhou, and S.~Y. Philip.
\newblock {\em Differential privacy and applications}, volume~69.
\newblock Springer, 2017.

\end{thebibliography}
\bibliographystyle{abbrv}

\end{document}